\documentclass[runningheads]{llncs}

\usepackage{xcolor}

\usepackage{graphicx}
\usepackage{comment}
\usepackage[many]{tcolorbox}
\newtcolorbox{mybox}{colback=blue!10}

\usepackage{parskip}

\begin{document}

\title{Securing the Future: Exploring Privacy Risks and Security Questions in Robotic Systems}
\titlerunning{Securing the Future: Exploring Privacy Risks
and Security Questions}
\author{Diba Afroze \orcidID{0009-0008-7859-7181}, Yazhou Tu \orcidID{0000-0001-7640-1829}, \and
Xiali Hei \orcidID{0000-0002-2438-5430} }
\authorrunning{D. Afroze et al.}
\institute{University of Louisiana at Lafayette, Lafayette, LA, 70504, USA }
\maketitle              % typeset the header of the contribution
\begin{abstract}
The integration of artificial intelligence, especially large language models in robotics, has led to rapid advancements in the field. We are now observing an unprecedented surge in the use of robots in our daily lives. The development and continual improvements of robots are moving at an astonishing pace. 
Although these remarkable improvements facilitate and enhance our lives, several security and privacy concerns have not been resolved yet. Therefore, it has become crucial to address the privacy and security threats of robotic systems while improving our experiences. In this paper, we aim to present existing applications and threats of robotics, anticipated future evolution, and the security and privacy issues they may imply. We present a series of open questions for researchers and practitioners to explore further. 

\end{abstract}

\section{Introduction}
The twenty-first century is witnessing an unprecedented increase in the evolution and utilization of robots. With the upcoming Industry 4.0 revolution, we are approaching the era of robotics \cite{industry4}. Currently, robotic systems play an important role, from performing medical procedures to serving as salespeople in shopping centers. Robots are now even replacing human companions. This remarkable growth, from a simple machine to an autonomous humanoid robot, has become possible because of the advancement of artificial intelligence, natural language processing, sensor technology, and processing power. 

To employ automation in work, different types of robots are used, designed to suit the specific nature of the work. We can categorize three general types of robots, i.e., \emph{Industrial Robots}, \emph{Service Robots}, and \emph{Specialized Robots} \cite{robottypes2}. Nowadays, these robots perform multipurpose applications seamlessly alongside humans in industries as well as at home. They handle heavy, mundane tasks for humans effortlessly. Additionally, they are becoming reliable in specialized tasks like healthcare assistance, surveillance, space exploration, rescue missions, etc. Robots are also helping as nurses or companions for older people. The vehicle industry is being revolutionized by the uprising of autonomous vehicles. All these advancements illustrate the prospect of reducing the gap between science fiction and reality.

As we embrace the help of robots in our daily lives, it may not be very long before these intelligent machines start to co-exist with us in society in every sector. Robotic help can undoubtedly simplify our lives, but it comes with potential privacy and security risks to our personal and social lives. Therefore, it is imperative to develop methods to prevent different kinds of privacy and security threats of robots to humans. Existing versions of robots are not free from threats, thereby indicating that future versions are unlikely to be different. There are several questions concerning privacy and security that a robot must answer before we may consider it to be safe to release in society. If we do not ensure that robots' mechanisms can answer these questions, we might have to reassess the deployment of robot among humans due to the inherent risk it poses to human life. In this paper, we explore a few of these questions.

In the following sections of this paper, we will address the growth of robotic advancement and several privacy and security-related questions that need our attention.

\section{Literature Review}
The proliferation of Robots is accelerating rapidly in our daily lives, and with it comes a rise in potential dangers. From the beginning of the use of robots, back in 1979, the first death induced by an industrial robot has been recorded \cite{acc}. After that, several deaths and injuries were caused by robots \cite{robotHazards}. Even though robot R\&D companies are trying to implement policies for secure interaction between humans and robots, new threats arise with the development of new robot technologies. 

Today, Robots are serving in many roles, such as security guards, salespeople, helping hands at home, nurses, etc. In emergency situations, humans might not follow the instructions of robots acting as security guards \cite{refuseRobotCommands}. An open question is: What would happen if people refuse to take commands from robots? Will the robot force humans or let them pass? Trust has not yet been fully established for robot services. People are concerned about their security; They are skeptical about letting unknown robots into their living spaces \cite{piggybacking}. Trust also depends on the appearance of robots; in some cases, people may feel threatened by humanoid robots that perform better than them at work \cite{humanoidRobots}. 

Robots are vulnerable to various forms of cyberattacks. Clark et al. present different cyber attack scenarios \cite{cyberattacks2}, for example, buffer overflow attacks to take control over companion robots, attacks on automated vehicles during firmware updates by pushing corrupted updates, hardware backdoor attacks on military drones, etc. Additionally, researchers show a comprehensive view of several cybersecurity issues such as malware, Trojan, replay attacks, fault injection, tampering attacks, etc. \cite{cyberattacks,cyberattack3,cybserattack4}.

Automated vehicles can be one of the targets of attackers. The attackers may use jamming, high-brightness Infrared LEDs, Digital Radio Frequency Memory (DRFM), etc. \cite{automatedVehicle}, to provide false navigation data. Additionally, autonomous vehicles are generally connected to users' smartphones. Sugawara et al. \cite{laserAttack} presented an audio injection attack on the voice-controlled smartphone system connected to automated Tesla and Ford cars. In addition, the classification system of autonomous vehicles is at risk of potential attack. The work in \cite{robustBackdoor,backdoor} demonstrated that a simple perturbation of the traffic signal could make the CNN classification model misidentify the signal. This attack poses significant security risks and can potentially cause chaos on roadways. Unmanned Automated Vehicles (UAVs), such as drones and rovers, are also in danger of being attacked. Dash et al. \cite{UAVs} demonstrated three attacks on UAVs protected by control invariants (CI) \cite{controlInvariants} and the extended Kalman filter (EKF) \cite{ekf}. The authors designed these attacks on UAVs by injecting minor false data into the control system which caused the automated vehicle to change its position and angular orientations, injecting time delays to make the UAV receive commands late, and lastly, injecting malicious code to switch the mode of the UAVs. In \cite{tu2018injected}, Tu et al. presented two attacks (i.e., Side Swing \cite{sideswing}, and DoS \cite{DoS}) to cyber-physical systems, and they manipulated two automatic self-balancing robots by spoofing embedded Micro Electro Mechanical Systems (MEMS) inertial sensors.

Telerobots \cite{telerobotics} come in handy in medical surgery, military operations, and rescue missions. In \cite{cyberSurgical,medicalTeleRObot}, the authors elaborated that telerobots are vulnerable to common cyber attacks such as viruses, worms, and malware. They also mentioned security threats such as command manipulation, denial of service, and communication loss. Recently, several medical centers have filed lawsuits against Intuitive Surgical, a surgical robot manufacturer, alleging that they were coerced into signing restrictive repair contracts, forcing them to buy new parts from the aforementioned company \cite{restrictiveContracts}. An operation had to be postponed due to the usage of third-party repair. This incident adds another dimension to the challenges of surgical robots.  
Shah et al. \cite{Sidechannelfingerprinting} demonstrated a successful side-channel attack- \emph{Fingeprint} on surgical robots. Besides, other potential side-channel attacks on robots are Radio-frequency attacks \cite{radioSidechannel} and cache-based attacks on automated vehicles \cite{cachesidechannel}.

Lutz et al. \cite{lutz2019privacy} observed robot usage from a different perspective, implying that social robots might affect the psychological and social privacy of human beings. Van et al. \cite{privacyCompromise} expressed their concern about whether we are compromising privacy in exchange for robotic services. \emph{The Guardian} reported \cite{barbie} about wifi-enabled Barbie dolls, which can be hacked and turned into a surveillance device to spy and collect information without anyone's knowledge. Robots are also becoming companions of humans, sometimes as caregivers. However, some authors are concerned about ethical issues. For example, the authors fear that companion robots might create a hallucinatory reality for some people \cite{companion}.

\section{Future Evolution and Security Questions}
Robots are evolving and becoming more intelligent, precise, and \emph{human-like}. Understandably, people are apprehensive about whether robots are going to be a threat to our lives, as depicted in science fiction movies. We are going to elaborate on some sectors for possible futuristic advancements in robots and the privacy and security questions that come with them.

\begin{itemize}
    \item \textbf{\emph{Cyber Security: }}
    Robots are now connected to wired and wireless networks for smooth data exchange and communication like any other device. However, robots have a lot of security issues, such as lack of authorization, authentication, secure network, tamper-resistant hardware, privacy, integrity, etc. \cite{cyberattacks}. 
Robotic networks and computer networks are different in nature; the same countermeasures in general computers may not work on robotics networks \cite{currentResearch}. Robotic Operating System (ROS) is also becoming popular among developers. Nevertheless, ROS is vulnerable to attacks such as DoS, DDoS attacks, malware, buffer overflow, malicious code injection attacks, etc. \cite{cyberattacks2}.  

Ransomware is another concern for robot users. In \cite{Akerbeltz}, Mayoral-Vilches et al. show a ransomware attack-\emph{Akerbeltz} on industrial robots, which locks and encrypts the robot from its vendor network. The attack was carried out by simply connecting a USB device to the robot or remotely accessing the adjacent network. Furthermore, another ransomware attack was demonstrated on a SoftBank Robotics NAO humanoid robot \cite{RansomwareBank}.  

\begin{mybox}
    \emph{\textbf{Open Question 1:}} Is there a way to identify security vulnerabilities early in robots? Is the robotic system software updated, or are security patches issued promptly? 
\end{mybox}

\vspace{\baselineskip} 
\item \textbf{\emph{IoT Connections:}} 
Robots are now becoming part of IoT and interconnecting with other devices. In homes, industries, and offices, it is common to connect robots with home assistants, smartphones, and TVs. Consider a scenario where an industrial robot integrates with other devices within a multi-purpose company. If an unauthorized user takes control of the robot, the whole system will be compromised. The attacker can take control of other devices and perform dangerous tasks. For example, this security breach may lead to injury, financial damage, and data theft. Thus, it is necessary to secure the additional mobile attack interface - robots. Another scenario is depicted by Amoozadeh et al. \cite{IotAutoVehicle}, where each vehicle receives beacon messages from the immediately preceding vehicle using the IEEE 802.11p protocol. The authors demonstrated security attacks (e.g., message falsification attack, spoofing attack, distributed DoS, Radio jamming, etc.), system-level attacks (e.g., hardware or software tempering), and privacy attacks (e.g., eavesdropping attack) on different layers of automated vehicle networks. A compromised network of vehicles can endanger passengers in all connected vehicles. Moreover, the attacker can evade privacy by leaking personal information such as vehicle identity, current vehicle position, speed, and acceleration.

\begin{mybox}
    \emph{\textbf{Open Question 2:}} How can the robot immediately detect and respond to a security breach? Can the robot alert the administrator about the intruder?
\end{mybox}

\vspace{\baselineskip} 
\item \textbf{\emph{Mutual Authentication: }}
Authentication has become one of the main concerns in robotics. Mutual authentication is necessary to establish secure communication between robots and humans. Several works have been done to authenticate users, such as face recognition, voice recognition \cite{currentResearch}, behavior-based recognition \cite{behaviour} etc. However, as we are employing an increasing number of robots in our work, the robots' identities need to be verified as well. Some delivery robots \cite{kiwibot,yandex,serverobot} use OTP (One-Time Password) or mobile applications on users' smartphones to authenticate to the user. But these methods are insufficient because they are susceptible to attacks \cite{OTPattack}. Adi et al. proposed an \emph{unclonable identity} for robots based on the work \cite{unclonable2}. This identity will be unique to human DNA. However, this process is complex, expensive, and not feasible for mass production. Later, Gavrilova et al. \cite{avatarBiometric} presented an idea to use biometric principles (e.g., physical and behavioral characteristics) to recognize and authenticate \emph{virtual} avatars. 

\begin{mybox}
    \emph{\textbf{Open Question 3:}} Is it possible to assign unique biometrics for robot authentication?
\end{mybox}

\vspace{\baselineskip} 
\item \textbf{\emph{Autonomous Robot: }}
The current generation of robots is not fully autonomous; they depend on pre-programmed commands. However, several initiatives are underway to extend the perimeter and allow robots to have autonomy to some extent, e.g., unmanned vehicles, Tesla bot \cite{Tesla}. 

Military services are also trying to utilize autonomous robots in war, spying, bomb defusal, and other dangerous jobs. However, the use of robots at war is a controversial topic, as it can violate international \emph{Humanitarian law} \cite{HumaniTarianLaw}. The question arises with the \emph{Robot at war}, what happens when an order contradicts the \emph{war robot}'s system. For example, if a robot receives an order to attack a house, the robot detects with sensors that the house is full of children. The order contradicts the robot's system in minimizing civilian casualties. Should the robot be allowed to have an awareness of these types of situations, or should the order override the robot's system \cite{roboticsinWar}?

\begin{mybox}
    \emph{\textbf{Open Question 4:}} What if autonomous robots start to make decisions or refuse orders that might cause harm to humans,  like kicking back a human who kicks it?
\end{mybox}

\vspace{\baselineskip} 
\item \textbf{\emph{Robot Learning: }}
Robot Learning \cite{robotLearning} is popular for teaching robots without programming every movement explicitly. Robots can learn from demonstrations, teleoperations, or observation \cite{robotLearning2}. Learning methods can be supervised, unsupervised, transfer learning, and reinforcement learning \cite{robotLearningTransfer}. The robots adapt their decisions as they perceive the environment or dataset. The attackers can intentionally manipulate the data during the learning process, such as injecting poisonous data into the training set, spoofing sensor data (e.g., camera, audio), or changing learning conditions. Due to these attacks, robots may learn unsolicited behaviors that can exhibit danger to their surroundings. For example, Yang et al. \cite{robotLearningAttack} demonstrated an adversarial attack on a reinforcement learning-based robot learning system where the attacker uses a pulse to generate random observations, degrading the learning performance.

\begin{mybox}
    \emph{\textbf{Open Question 5:}} How can anomalies in robot training data be discovered and addressed so that the robot does not learn and perpetuate dangerous behavior?
\end{mybox}

\vspace{\baselineskip}
\item \textbf{\emph{Integration with ChatGPT: }}
Robots are expected to undergo revolutionary changes using ChatGPT, especially ChatGPT-4. We have seen some proposed frameworks \cite{chatgptPrompt,robotgpt} in recent times. Vemprala et al. \cite{chatgptPrompt} suggested using a ChatGPT prompt to write code automatically for non-technical users to make the robot perform a certain task. In one scenario, the user asks the robot to cook an omelet and serves it to the user's grandfather. Recently, Google DeepMind introduced Robotic Transformer 2 (RT-2), a novel vision-language-action (VLA) model that learns from web-scale datasets \cite{GoogleDeepMind}. This model is built on the same tech as ChatGPT; It can interpret these data as plain language instruction and execute it \cite{WallE}. 

\begin{mybox}
     \emph{\textbf{Open Question 6:}} If ChatGPT can be successfully implemented on robots, what if robots can write code and modify themselves in an unwanted way?
\end{mybox}

\vspace{\baselineskip}
\item \textbf{\emph{Access Control: }}
Certain robots (e.g., service robots in our homes) continuously surveil us as part of their functions. These robots have access to our personal data; they can take pictures and videos, and monitor our locations. Nonetheless, if the vendor of these robots unethically grants access to the robots' system during manufacturing and takes advantage of our confidential data, it can pose significant privacy and security risks. For example, unauthorized users can collect passwords and credit card information by simply taking photos or videos through robots when the user is entering the data.

\begin{mybox}
     \emph{\textbf{Open Question 7:}} How can we effectively incorporate access control in robots to protect the security and privacy of the end users?
\end{mybox}

\vspace{\baselineskip}
\item \textbf{\emph{Trolley Problem in Robotics: }}
Imagine a scenario where a person is watching a runaway trolley heading towards a track where five people are standing, and if nothing is done, these people will \emph{certainly} die. There is another track where he can divert the trolley, but there is another person standing on it who will be killed. Here arises the ethical dilemma of whether killing one person is okay instead of killing five people. As robots become more involved in society, they will inevitably encounter many ethical dilemmas in decision-making. So, it is essential to solve the trolley problem to mitigate any risks that an action of the robot may pose.

\begin{mybox}
    \emph{\textbf{Open Question 8:}} What would be the robot’s reaction during a ‘Trolley Problem’ \cite{trolley} scenario?
\end{mybox}
\end{itemize}

\section{Conclusion}
The widespread adoption of robots signals the imminent revolution of robotics technology. It may not be very long before we generalize the idea of coexisting with robots. We must be prepared for the privacy and security risks to embrace this transition fully. Robotic systems are made of different subsystems and subcomponents. Securing the subcomponents is necessary but not sufficient for protecting the whole system. This is because components are integrated with one another and therefore, exhibit complex and subtle dependencies and interactions \cite{satc2}. We need to enforce a robotics framework and a universal policy for developing or changing any robots. Such a comprehensive measure will ensure that robots and their manufacturer follow the standard user safety practice.
European Commission has created a \emph{voluntary} code of ethics and standards for manufacturers and users of robotics technology \cite{EC}. IEEE undertakes a global initiative-\emph{The IEEE Global Initiative on Ethics of Autonomous and Intelligent Systems}, which aims to ensure that the involved persons prioritize ethical consideration and benefits of humankind \cite{ieeeglobal}. However, as these policies are not enforced as obligatory, the concerns still prevail.

\bibliographystyle{splncs04}
\bibliography{mybibliography}

\begin{thebibliography}{10}
\providecommand{\url}[1]{\texttt{#1}}
\providecommand{\urlprefix}{URL }
\providecommand{\doi}[1]{https://doi.org/#1}

\bibitem{unclonable2}
Adi, W.: Clone-resistant dna-like secured dynamic identity. In: 2008 Bio-inspired, Learning and Intelligent Systems for Security. pp. 148--153 (2008). \doi{10.1109/BLISS.2008.33}

\bibitem{refuseRobotCommands}
Agrawal, S., Williams, M.A.: Robot authority and human obedience: A study of human behaviour using a robot security guard. In: Proceedings of the companion of the 2017 ACM/IEEE international conference on human-robot interaction. pp. 57--58 (2017)

\bibitem{behaviour}
Almohamade, S.S., Clark, J.A., Law, J.: Behaviour-based biometrics for continuous user authentication to industrial collaborative robots. In: Innovative Security Solutions for Information Technology and Communications: 13th International Conference, SecITC 2020, Bucharest, Romania, November 19--20, 2020, Revised Selected Papers 13. pp. 185--197. Springer (2021)

\bibitem{IotAutoVehicle}
Amoozadeh, M., Raghuramu, A., Chuah, C.N., Ghosal, D., Zhang, H.M., Rowe, J., Levitt, K.: Security vulnerabilities of connected vehicle streams and their impact on cooperative driving. IEEE Communications Magazine  \textbf{53}(6),  126--132 (2015)

\bibitem{cyberSurgical}
Bernadotte, A.: Cyber security for surgical remote intelligent robotic systems. In: 2023 9th International Conference on Automation, Robotics and Applications (ICARA). pp. 65--69 (2023). \doi{10.1109/ICARA56516.2023.10126050}

\bibitem{companion}
Bisconti~Lucidi, P., Nardi, D.: Companion robots: the hallucinatory danger of human-robot interactions. In: Proceedings of the 2018 AAAI/ACM Conference on AI, Ethics, and Society. pp. 17--22 (2018)

\bibitem{medicalTeleRObot}
Bonaci, T., Herron, J., Yusuf, T., Yan, J., Kohno, T., Chizeck, H.J.: To make a robot secure: An experimental analysis of cyber security threats against teleoperated surgical robots. arXiv preprint arXiv:1504.04339  (2015)

\bibitem{piggybacking}
Booth, S., Tompkin, J., Pfister, H., Waldo, J., Gajos, K., Nagpal, R.: Piggybacking robots: Human-robot overtrust in university dormitory security. In: Proceedings of the 2017 ACM/IEEE International Conference on Human-Robot Interaction. pp. 426--434 (2017)

\bibitem{ekf}
Bristeau, P.J., Dorveaux, E., Vissi{\`e}re, D., Petit, N.: Hardware and software architecture for state estimation on an experimental low-cost small-scaled helicopter. Control Engineering Practice  \textbf{18}(7),  733--746 (2010)

\bibitem{controlInvariants}
Choi, H., Lee, W.C., Aafer, Y., Fei, F., Tu, Z., Zhang, X., Xu, D., Deng, X.: Detecting attacks against robotic vehicles: A control invariant approach. In: Proceedings of the 2018 ACM SIGSAC Conference on Computer and Communications Security. pp. 801--816 (2018)

\bibitem{cyberattacks2}
Clark, G.W., Doran, M.V., Andel, T.R.: Cybersecurity issues in robotics. In: 2017 IEEE conference on cognitive and computational aspects of situation management (CogSIMA). pp.~1--5. IEEE (2017)

\bibitem{robotLearning}
Connell, J.H., Mahadevan, S.: Robot learning, vol.~233. Springer Science \& Business Media (2012)

\bibitem{UAVs}
Dash, P., Karimibiuki, M., Pattabiraman, K.: Out of control: stealthy attacks against robotic vehicles protected by control-based techniques. In: Proceedings of the 35th Annual Computer Security Applications Conference. pp. 660--672 (2019)

\bibitem{WallE}
EDWARDS, B.: {Google’s RT-2 AI model brings us one step closer to WALL-E}. \url{https://arstechnica.com/information-technology/2023/07/googles-rt-2-ai-model-brings-us-one-step-closer-to-wall-e/} (July 28, 2023)

\bibitem{robustBackdoor}
Eykholt, K., Evtimov, I., Fernandes, E., Li, B., Rahmati, A., Xiao, C., Prakash, A., Kohno, T., Song, D.: Robust physical-world attacks on deep learning visual classification. In: Proceedings of the IEEE conference on computer vision and pattern recognition. pp. 1625--1634 (2018)

\bibitem{avatarBiometric}
Gavrilova, M.L., Yampolskiy, R.V.: Applying biometric principles to avatar recognition. In: 2010 International Conference on Cyberworlds. pp. 179--186 (2010). \doi{10.1109/CW.2010.36}

\bibitem{privacyCompromise}
van den Hoven~van Genderen, R.: Privacy and data protection in the age of pervasive technologies in ai and robotics. European Data Protection Law Review  \textbf{3},  338--352 (01 2017). \doi{10.21552/edpl/2017/3/8}

\bibitem{barbie}
Gibbs, S.: {Hackers can hijack Wi-Fi Hello Barbie to spy on your children)}. \url{ https://www.theguardian.com/technology/2015/nov/26/hackers-can-hijack-wi-fi-hello-barbie-to-spy-on-your-children} (november, 2015)

\bibitem{robotgpt}
He, H.M.: Robot{GPT}: From chat{GPT} to robot intelligence (2023), \url{https://openreview.net/forum?id=wWe\_OqpCcU8}

\bibitem{ieeeglobal}
IEEE: {The IEEE Global Initiative on Ethics of Autonomous and Intelligent Systems}. \url{https://standards.ieee.org/wp-content/uploads/import/documents/other/ec\_about\_us.pdf} (2017)

\bibitem{DoS}
Injected, Demos, D.: {DoS attacks on a self-balancing robot (accelerometer)}. \url{ https://youtu.be/yDz8y\_ht3Xg} (February, 2018)

\bibitem{sideswing}
Injected, Demos, D.: {Side-Swing attacks on a self-balancing robot}. \url{https://youtu.be/oy3B1X41u5s} (February, 2018)

\bibitem{robottypes2}
{International Federation of Robotics (IFR)}: {Service Robots as Defined by ISO 8373}. \url{https://ifr.org/service-robots}

\bibitem{trolley}
Kamm, F.M.: The Trolley Problem Mysteries. Oxford University Press (2015)

\bibitem{robotHazards}
Kirschgens, L.A., Ugarte, I.Z., Uriarte, E.G., Rosas, A.M., Vilches, V.M.: Robot hazards: from safety to security. arXiv preprint arXiv:1806.06681  (2018)

\bibitem{kiwibot}
Kiwibot: {Kiwibot}. \url{https://www.kiwibot.com/}

\bibitem{robotLearning2}
Kroemer, O., Niekum, S., Konidaris, G.: A review of robot learning for manipulation: Challenges, representations, and algorithms. The Journal of Machine Learning Research  \textbf{22}(1),  1395--1476 (2021)

\bibitem{cybserattack4}
Lacava, G., Marotta, A., Martinelli, F., Saracino, A., La~Marra, A., Gil-Uriarte, E., Vilches, V.M.: Cybsersecurity issues in robotics. J. Wirel. Mob. Networks Ubiquitous Comput. Dependable Appl.  \textbf{12}(3),  1--28 (2021)

\bibitem{RansomwareBank}
Larson, S.: {Ransomware experiment shows the dangers of hacking robots}. \url{https://money.cnn.com/2018/03/09/technology/robots-ransomware/index.html} (March 9, 2018)

\bibitem{roboticsinWar}
Lin, P., Bekey, G.A., Abney, K.: Robots in war: issues of risk and ethics  (2009)

\bibitem{backdoor}
Liu, Y., Ma, X., Bailey, J., Lu, F.: Reflection backdoor: A natural backdoor attack on deep neural networks. In: Computer Vision--ECCV 2020: 16th European Conference, Glasgow, UK, August 23--28, 2020, Proceedings, Part X 16. pp. 182--199. Springer (2020)

\bibitem{cachesidechannel}
Luo, M., Myers, A.C., Suh, G.E.: Stealthy tracking of autonomous vehicles with cache side channels. In: 29th USENIX Security Symposium (USENIX Security 20). pp. 859--876 (2020)

\bibitem{lutz2019privacy}
Lutz, C., Sch{\"o}ttler, M., Hoffmann, C.P.: The privacy implications of social robots: Scoping review and expert interviews. Mobile Media \& Communication  \textbf{7}(3),  412--434 (2019)

\bibitem{Akerbeltz}
Mayoral-Vilches, V., Carbajo, U.A., Gil-Uriarte, E.: Industrial robot ransomware: Akerbeltz. In: 2020 Fourth IEEE International Conference on Robotic Computing (IRC). pp. 432--435 (2020). \doi{10.1109/IRC.2020.00080}

\bibitem{satc2}
McDaniel, P., Koushanfar, F.: Secure and trustworthy computing 2.0 vision statement. arXiv preprint arXiv:2308.00623  (2023)

\bibitem{OTPattack}
Mulliner, C., Borgaonkar, R., Stewin, P., Seifert, J.P.: Sms-based one-time passwords: Attacks and defense: (short paper). In: Detection of Intrusions and Malware, and Vulnerability Assessment: 10th International Conference, DIMVA 2013, Berlin, Germany, July 18-19, 2013. Proceedings 10. pp. 150--159. Springer (2013)

\bibitem{EC}
Nevejans, N.: {EUROPEAN CIVIL LAW RULES IN ROBOTICS}. \url{http://www.europarl.europa.eu/committees/fr/supporting-analyses-search.html} (October, 2016)

\bibitem{telerobotics}
Niemeyer, G., Preusche, C., Stramigioli, S., Lee, D.: Telerobotics. Springer handbook of robotics pp. 1085--1108 (2016)

\bibitem{industry4}
Othman, F., Bahrin, M., Azli, N., et~al.: Industry 4.0: A review on industrial automation and robotic. J Teknol  \textbf{78}(6-13),  137--143 (2016)

\bibitem{automatedVehicle}
Petit, J., Shladover, S.E.: Potential cyberattacks on automated vehicles. IEEE Transactions on Intelligent Transportation Systems  \textbf{16}(2),  546--556 (2015). \doi{10.1109/TITS.2014.2342271}

\bibitem{robotLearningTransfer}
Ranaweera, M., Mahmoud, Q.H.: Virtual to real-world transfer learning: A systematic review. Electronics  \textbf{10}(12), ~1491 (2021)

\bibitem{restrictiveContracts}
REUTER, E.: {Hospitals sue surgical robot maker, saying it forced them into restrictive contracts}. \url{https://medcitynews.com/2021/07/hospitals-sue-surgical-robot-maker-saying-it-forced-them-into-restrictive-contracts/} (July 14, 2021)

\bibitem{serverobot}
Serve: {Serve Robotics Becomes First Autonomous Vehicle Company to Commercially Launch Level 4 Self-Driving Robots}. \url{https://www.serverobotics.com/level-4-autonomy}

\bibitem{Sidechannelfingerprinting}
Shah, R., Ahmed, M., Nagaraja, S.: Fingerprinting robot movements via acoustic side channel. arXiv preprint arXiv:2209.10240  (2022)

\bibitem{radioSidechannel}
Shah, R., Ahmed, M., Nagaraja, S.: Reconstructing robot operations via radio-frequency side-channel. arXiv preprint arXiv:2209.10179  (2022)

\bibitem{laserAttack}
Sugawara, T., Cyr, B., Rampazzi, S., Genkin, D., Fu, K.: Light commands: {Laser-Based} audio injection attacks on {Voice-Controllable} systems. In: 29th USENIX Security Symposium (USENIX Security 20). pp. 2631--2648. USENIX Association (Aug 2020), \url{https://www.usenix.org/conference/usenixsecurity20/presentation/sugawara}

\bibitem{HumaniTarianLaw}
Szpak, A.: Legality of use and challenges of new technologies in warfare – the use of autonomous weapons in contemporary or future wars. European Review  \textbf{28}(1),  118–131 (2020). \doi{10.1017/S1062798719000310}

\bibitem{yandex}
Team, Y.S.D.: {The story behind the creation of Yandex’s delivery robot}. \url{https://medium.com/yandex-self-driving-car/the-story-behind-the-creation-of-yandexs-delivery-robot-e07017940589} (December 16, 2021)

\bibitem{Tesla}
Tesla: {Tesla Bot Update}. \url{https://www.youtube.com/watch?v=XiQkeWOFwmk} (May 2023)

\bibitem{tu2018injected}
Tu, Y., Lin, Z., Lee, I., Hei, X.: Injected and delivered: Fabricating implicit control over actuation systems by spoofing inertial sensors. In: 27th USENIX Security Symposium (USENIX Security 18). pp. 1545--1562 (2018)

\bibitem{chatgptPrompt}
Vemprala, S., Bonatti, R., Bucker, A., Kapoor, A.: Chatgpt for robotics: Design principles and model abilities. Microsoft Auton. Syst. Robot. Res  \textbf{2}, ~20 (2023)

\bibitem{currentResearch}
Wang, T.M., Tao, Y., Liu, H.: Current researches and future development trend of intelligent robot: A review. International Journal of Automation and Computing  \textbf{15}(5),  525--546 (2018)

\bibitem{acc}
Winfield, A.F., Winkle, K., Webb, H., Lyngs, U., Jirotka, M., Macrae, C.: Robot accident investigation: a case study in responsible robotics. Software engineering for robotics pp. 165--187 (2021)

\bibitem{cyberattacks}
Yaacoub, J.P.A., Noura, H.N., Salman, O., Chehab, A.: Robotics cyber security: Vulnerabilities, attacks, countermeasures, and recommendations. International Journal of Information Security pp. 1--44 (2022)

\bibitem{robotLearningAttack}
Yang, C.H.H., Qi, J., Chen, P.Y., Ouyang, Y., Hung, I.T.D., Lee, C.H., Ma, X.: Enhanced adversarial strategically-timed attacks against deep reinforcement learning. In: ICASSP 2020-2020 IEEE International Conference on Acoustics, Speech and Signal Processing (ICASSP). pp. 3407--3411. IEEE (2020)

\bibitem{GoogleDeepMind}
Yevgen~Chebotar, T.Y.: {RT-2: New model translates vision and language into action}. \url{https://www.deepmind.com/blog/rt-2-new-model-translates-vision-and-language-into-action} (July 28, 2023)

\bibitem{humanoidRobots}
Yogeeswaran, K., Z{\l}otowski, J., Livingstone, M., Bartneck, C., Sumioka, H., Ishiguro, H.: The interactive effects of robot anthropomorphism and robot ability on perceived threat and support for robotics research. Journal of Human-Robot Interaction  \textbf{5}(2),  29--47 (2016)

\bibitem{cyberattack3}
Zhu, Q., Rass, S., Dieber, B., Vilches, V.M., et~al.: Cybersecurity in robotics: Challenges, quantitative modeling, and practice. Foundations and Trends{\textregistered} in Robotics  \textbf{9}(1),  1--129 (2021)

\end{thebibliography}

\end{document}